\documentclass{article}
\usepackage{amsmath,amssymb,amsfonts}
\usepackage{algorithmic}
\usepackage{graphicx}
\usepackage{textcomp}
\usepackage{multirow}
\usepackage{booktabs}
\usepackage{colortbl}
\usepackage{arydshln}

\usepackage{subcaption}
\usepackage[utf8]{inputenc}
\usepackage[margin=1.25in]{geometry}
\usepackage{authblk}
\usepackage{setspace}

\usepackage{float}
\definecolor{subsectioncolor}{rgb}{0,0.541,0.855}
\usepackage[style=ieee, 
citestyle=numeric-comp,
sorting=none]{biblatex}
\title{Improving Prediction of Cognitive Performance using Deep Neural Networks in Sparse Data}

\author[1*]{Sharath Koorathota}
\author[2]{Arunesh Mittal}
\author[3]{Richard Sloan}
\author[1,2,4]{Paul Sajda}

\affil[1]{Department of Biomedical Engineering, Columbia University, New York,  USA.}
\affil[2]{Department of Electrical Engineering, Columbia University, New York,  USA.}
\affil[3]{Department of Psychiatry, Columbia University Medical Center, New York, USA.}
\affil[4]{Data Science Institute, Columbia University, New York, USA.}
\affil[*]{Corresponding author. Email: sk4172@columbia.edu}

\date{}

\begin{document}

\maketitle

\begin{abstract}
Cognition in midlife is an important predictor of age-related mental decline and statistical models that predict cognitive performance can be useful for predicting decline. However, existing models struggle to capture complex relationships between physical, sociodemographic, psychological and mental health factors that effect cognition. Using data from an observational, cohort study, Midlife in the United States (MIDUS), we modeled a large number of variables to predict executive function and episodic memory measures. We used cross-sectional and longitudinal outcomes with varying sparsity, or amount of missing data. Deep neural network (DNN) models consistently ranked highest in all of the cognitive performance prediction tasks, as assessed with root mean squared error (RMSE) on out-of-sample data. RMSE differences between DNN and other model types were statistically significant (T(8) = -3.70; p $<$ 0.05).  The interaction effect between model type and sparsity was significant ($F_9$=59.20; p $<$ 0.01), indicating the success of DNNs can partly be attributed to their robustness and ability to model hierarchical relationships between health-related factors. Our findings underscore the potential of neural networks to model clinical datasets and allow better understanding of factors that lead to cognitive decline.
\end{abstract}
\section{Introduction}

Cognitive performance is a predictor of a variety of age-related clinical conditions. For example, substantial research has linked measures of cognitive performance and its decline to conditions such as attention-deficit/hyperactivity disorder (ADHD), mild cognitive impairment (MCI), and Alzheimer’s disease (AD) \cite{soderlund2007listen, der2007effect, baum1993cognitive, sabbagh2020early, doraiswamy1997alzheimer}. Furthermore, interventions enhancing cognitive performance have had measurable effects on progression of MCI \cite{olchik2013memory, devenney2017effects}.  Improving prediction of cognitive performance has the potential to impact the diagnosis and treatment of mental illness.

Cognitive performance is affected by a host of factors. A complex system of physical, psychosocial and emotional factors affect performance in patient and healthy populations  \cite{Tompkins1990PredictingAdolescents}. For example, better executive function and episodic memory have been associated with demographic variables such as sex \cite{Hughes2018ChangeStudy}, recovery of vagal tone from cognitive challenge \cite{Hansen2003} and higher educational attainment \cite{Hughes2018ChangeStudy}. The majority of past approaches to understanding factors that are associated with cognitive performance have used linear models, such as regression, where the relationship between variables or factors can be reported through beta weights \cite{Raslear2011PredictingRisk, OSullivan2009HippocampalCADASIL, Graham2012PersonalityAble, Hughes2018ChangeStudy}, with more recent work using interpretable, non-linear models such as random forests (RF) \cite{Casanova2020InvestigatingLearning}. 

A challenge in predicting cognitive performance in clinical populations is the trade-off in the size of a dataset and the quality. For large, national, studies with a broad aim, data may be collected at multiple sites, in-lab (e.g., biological specimen data) or remotely (e.g., phone questionnaires). Statistical models in this case may suffer from overfitting the data, driven by large number of missing data \cite{Ilin2010} or sparse variables, where only a small proportion of the participant pool has measured values \cite{Hawthorne2005, Garcia-Laencina2015, Fielding2009}. Furthermore, model fitting and assessment in this area of research is often performed on the same samples. While this approach makes sense in order to understand associations between variables and their strength in predicting cognitive performance, these methods are not ideal for improving the robustness and generalizability of prediction itself.

The primary purpose of this study is to improve cross-sectional and longitudinal prediction of cognitive performance. To this end, we use aggregated data from the Midlife in the United States (MIDUS) study. Our innovations are twofold. First, we introduce non-linear modeling frameworks which yield lower prediction errors across multiple cognitive outcome measures compared to traditional, statistical models. Specifically, we show deep neural networks (DNNs) outperform other models through learning the complex relationships of factors in a heterogeneous, sparse dataspace. Secondly, we present a method for interpreting hierarchical relationships between clinical factors using vectorized representations, or embeddings, of groups of related variables. Together, our results highlight the advantages of DNNs in the context of modeling large, mixed, and sparse clinical datasets. 

\section{Methods}

\subsection{Dataset}
The MIDUS-AGG dataset was aggregated from the MIDUS national study \cite{Brim2004TheOverview, Ryff2017Midlife2004-2006}. The overall goal of MIDUS is to investigate the factors that influence physical and mental health as people age. The current study focused on the participants who completed the cognitive assessment from the second and third waves of the longitudinal study. The sample was recruited in 1995–1996 through random digit dialing of U.S. households having at least one telephone in the contiguous 48 states, stratified by age with an oversample of those between 40 and 60 years of age. 

A second wave (MIDUS 2, M2) of data collection took place in 2004-2005, in 4955 participants from M1 (53.8\% female; 90.1\% white) aged 32 to 84 years (M=55.36; SD=12.40), with a mean education level of 14.24 years (SD=2.60) \cite{Radler2010WhoWell-Being}. A third wave of data collection (MIDUS 3, M3) was conducted on average 9 years (SD=.53) after M2, with 76.9\% retention of eligible M1 participants (N=3294; 55.3\% female; 90.4\% white). Whereas in M1, only survey data were collected, in M2 ad M3, data were collected in daily stressor, cognitive, biomarker, and neuroimaging projects.  Participants ranged in age from 42 to 92 years (M=64.30; SD=11.20) at M3. Typical of longitudinal studies, MIDUS 3 participants showed some differences on MIDUS 2 variables compared with those who dropped out of the study, with dropouts performing significantly more poorly on all cognitive tests at MIDUS 2 \cite{Hughes2018ChangeStudy}.

\subsubsection{Cognitive Measures}
Cognitive performance was assessed using the
Brief Test of Adult Cognition by Telephone (BTACT), designed to be sensitive to changes in cognitive function \cite{Tun2006TelephoneTelephone, Lachman2014MonitoringTelephone}. The BTACT included assessed word-list free recalls, digits backward span tasks, verbal fluency tasks, number series pattern tasks and the Stop and Go Switch Task.  In addition, composite scores for executive function (EF) and episodic memory (EM) were computed.   

In the present study, we used machine learning approaches applied to data from all MIDUS projects to predict EF, EM and COMP at M2 and M3 and change scores ($\Delta$) from M2 to M3 in 9 separate prediction tasks. We defined sparsity simply as the count of missing, invalid or inapplicable data in each prediction task. Selected participant demographics, variable counts, and
sparsity of the datasets used for prediction of each outcome are listed in Table~\ref{descriptive}. 

\begin{table*}[ht]
\centering
\resizebox{\textwidth}{!}{
\begin{tabular}{@{}lllllllllll@{}}
\toprule
                  & \multicolumn{4}{c}{n}                                                     & \multicolumn{4}{c}{Score (SD)}                                            & \multicolumn{1}{c}{\multirow{3}{*}{Variable Count}} & \multicolumn{1}{c}{\multirow{3}{*}{Sparsity}} \\ \cmidrule(r){2-5}\cmidrule(r){6-9}\multicolumn{1}{c}{M2 Age}           & \multicolumn{2}{c}{\textless{}55} & \multicolumn{2}{c}{\textgreater{}=55} & \multicolumn{2}{c}{\textless{}55} & \multicolumn{2}{c}{\textgreater{}=55} & \multicolumn{1}{c}{}                                 & \multicolumn{1}{c}{}                          \\\cmidrule(r){1-9}
\multicolumn{1}{c}{Sex}               & Female           & Male           & Female             & Male             & Female          & Male            & Female            & Male              & \multicolumn{1}{c}{}                                 & \multicolumn{1}{c}{}                          \\ \cmidrule(l){1-11} 
\multicolumn{1}{c}{Prediction Task}               &                  &                &                    &                  &                 &                 &                   &                   &                                                      &                                               \\\cmidrule(l){1-1}
\multicolumn{1}{c}{COMP(M2)}          & 257              & 194            & 243                & 187              & 80.12 (16.84)   & 82.48 (16.62)   & 70.07 (15.74)     & 70.02 (15.55)     & 8542                                                 & 3459284                                       \\ \hdashline
\multicolumn{1}{c}{EF(M2)}            & 259              & 191            & 236                & 188              & 66.16 (15.01)   & 70.48 (14.35)   & 57.1 (13.28)      & 60.98 (14.04)     & 8542                                                 & 3445748                                       \\\hdashline
\multicolumn{1}{c}{EM(M2)}            & 249              & 186            & 242                & 184              & 12.62 (3.75)    & 10.97 (3.86)    & 11.43 (4.41)      & 8.47 (3.29)       & 8542                                                 & 3475186                                       \\\hdashline
\multicolumn{1}{c}{COMP(M3)}          & 197              & 145            & 159                & 140              & 81.47 (15.41)   & 82.1 (17.37)    & 68.31 (15.74)     & 69.41 (14.75)     & 9831                                                 & 2660931                                       \\\hdashline
\multicolumn{1}{c}{$\Delta$ COMP(M3)} & 188              & 142            & 153                & 133              & -2.18 (9.75)    & -2.92 (9.91)    & -5.15 (9.22)      & -3.86 (8.42)      & 9831                                                 & 2560627                                       \\\hdashline
\multicolumn{1}{c}{EF(M3)}            & 193              & 147            & 166                & 144              & 65.76 (13.66)   & 69.74 (15.36)   & 54.81 (13.83)     & 59.57 (13.2)      & 9831                                                 & 2691342                                       \\\hdashline
\multicolumn{1}{c}{$\Delta$ EF(M3)}   & 188              & 149            & 160                & 141              & -3.4 (8.71)     & -2.7 (9.1)      & -4.35 (7.79)      & -3.28 (6.87)      & 9831                                                 & 2627376                                       \\\hdashline
\multicolumn{1}{c}{EM(M3)}            & 194              & 147            & 162                & 143              & 13.28 (4.23)    & 10.69 (3.74)    & 11.18 (4.58)      & 8.31 (3.31)       & 9831                                                 & 2697842                                       \\\hdashline
\multicolumn{1}{c}{$\Delta$ EM(M3)}   & 173              & 146            & 158                & 138              & 0.14 (3.94)     & -0.72 (3.72)    & -1.09 (3.84)      & -0.97 (3.25)      & 9831                                                 & 2561928                                       \\
\bottomrule
\end{tabular}}
\caption{Description of the 9 prediction tasks: their outcome, cognitive measure, and the input data breakdown. Cognitive outcome measures included BTACT composite score (COMP), episodic memory score (EM), executive function score (EF). Note that COMP is comprised of EF and EM scores. Both MIDUS 2 (M2) and MIDUS 3 (M3) timepoints were predicted, along with change from M2 to M3 ($\Delta$). Only participants who complete all M2 projects and relevant M3 projects (Survey Measures, Cognitive Measures) were included in the study. Predictor variable count varied with task, and sparsity is simply quantified as the count of missing, inapplicable or invalid data points.}
\label{descriptive}
\end{table*}

\subsubsection{Survey Measures}
Survey data included sociodemographic, physical and mental health history. For example, a series of questions addressed alcohol use, including history of use, regularity of use, attempts to quit, and how the use of those substances affected respondents' physical and mental well-being. Questions included a mix of numerical, interval, ordinal and scaled data. Survey measures also asked about their perceived well-being compared to peers, exercise and diet habits, experience with menopause, and past therapies attempted. Demographic questions included work histories, presence or absence of their parents in childhood, religion, history of physical abuse, and the quality of their relationships to those around them. Respondents also were asked about their significant others' occupations, and about familial relationships.

\subsubsection{Daily Stress Measures}
Daily stress measures also were mixed in variable types, and included sociodemographic, health status, personality characteristics, and genetic factors related to exposure, coping and reactivity to day-to-day life stressors. These were obtained using a daily telephone interview, with participants, on average, completing 7.2 out of 8 days of surveys. Participants were asked about frequency, content and severity of daily stressors and how they compared to levels of stress in the past 10 years. They were also asked, through open-ended questions, about personality characteristics that may  affect their response to stress.

\subsubsection{Biomarker Measures}
Biomarker measures from the MIDUS dataset related to non-disease specific, biological data collection to understand how the functioning of various biological systems relate to behavioral and psychosocial factors. Specifically, biomarkers assessed functionality of the hypothalamic-pituitary-adrenal axis, autonomic nervous system, immune system, cardiovascular system, musculoskeletal system, antioxidants, and metabolic processes. Data related to biomarkers were collected through specimens, clinical assessments, self-reports and in-lab protocols. Notably, self-reported sleep assessments were also collected as a part of the Biomarker project.

\subsubsection{Data Aggregation}
We used two main approaches to aggregating MIDUS data for the purposes of this study. As a first step, we collected all M2 data available and all M3 data from projects that were complete (Survey Measures, Cognitive Measures; other projects are ongoing) from ICPSR, an online data archive \cite{Ryff2017Midlife2004-2006}. We then hand-annotated variables to include or exclude from analysis. Variables that were unique identifiers of participants were excluded (e.g., the date of their phone calls), are specific to the MIDUS study with potential to not be replicable, or had zero variance.  

We used hierarchical metadata from the original dataset to assign factor-level groups (Supplementary Table~\ref{tab: var-groupings}). Missing metadata from the MIDUS dataset were annotated and a total of 10,134 usable variables from M2 and M3 were isolated for processing.

\subsection{Data Processing}

Only participants who completed the cognitive, survey, daily stress and biomarker projects were included in the analyses.  Numerical data were normalized and categorical data were one-hot encoded prior to training.

\subsection{Prediction Approach}

\begin{figure}
    \centering
    \includegraphics[width=0.5\textwidth]{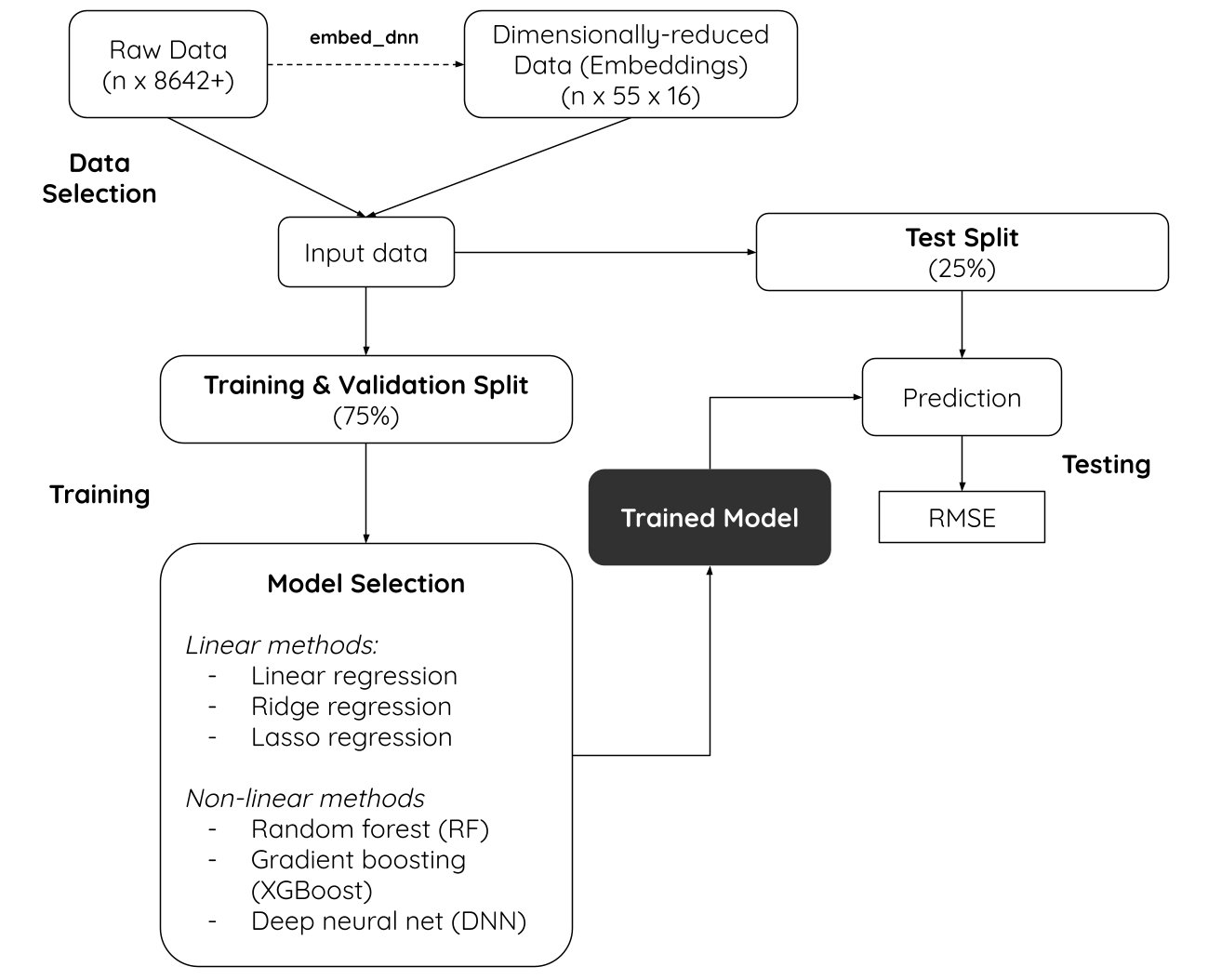}
    \caption{Overview of methods in the present study. We used the original MIDUS-AGG with over 10,134 variables and a variable number of participants, depending on the prediction task. Separately, we used an embedding representation of data as input to the DNN. In embeddings, each of the 55 factors were represented by a 16-dimensional vector, learned during training. Training utilized one of 6 models (linear or non-linear methods). Model performance was assessed using mean RMSE values from 10 runs, on random splits of the test data.}
    \label{fig:methods-overview}
\end{figure}

We report two, separate approaches to predicting cognitive performance (Figure~\ref{fig:methods-overview}). The first approach used the full dataset and the second approach used lower-dimensional representations. We expected that the sparseness and high dimensionality of MIDUS-AGG would make it difficult to avoid common issues such as overfitting \cite{Hawkins2004TheOverfitting} and curse-of-dimensionality \cite{Friedman1997OnCurse-of-dimensionality}. 

\subsubsection{Baseline Models}

To establish baseline comparisons, we selected models that are typically used for cognitive performance prediction, through a literature search of common approaches. We identified three model classes that have been used most often by researchers: linear regression methods, regression with regularization, and tree-based machine learning frameworks. We selected the following baseline, off-the-shelf \cite{Pedregosa2011Scikit-learnPython, Nori2019InterpretML:Interpretability} models for fitting (Figure~\ref{fig:methods-overview}):
\begin{enumerate}
    \item linear regression with no regularization,
    \item ridge regression with L2 regularization,
    \item lasso regression with L1 regularization,
    \item tree-based, RF,
    \item gradient boosting (XGBoost)
\end{enumerate}

RF and XGBoost are machine learning approaches that have yielded encouraging, interpretable results in clinical data \cite{Caruana2015IntelligibleHealthCare}. 

\subsubsection{Deep Neural Network}

\begin{figure}
    \centering
    \includegraphics[width=1\textwidth]{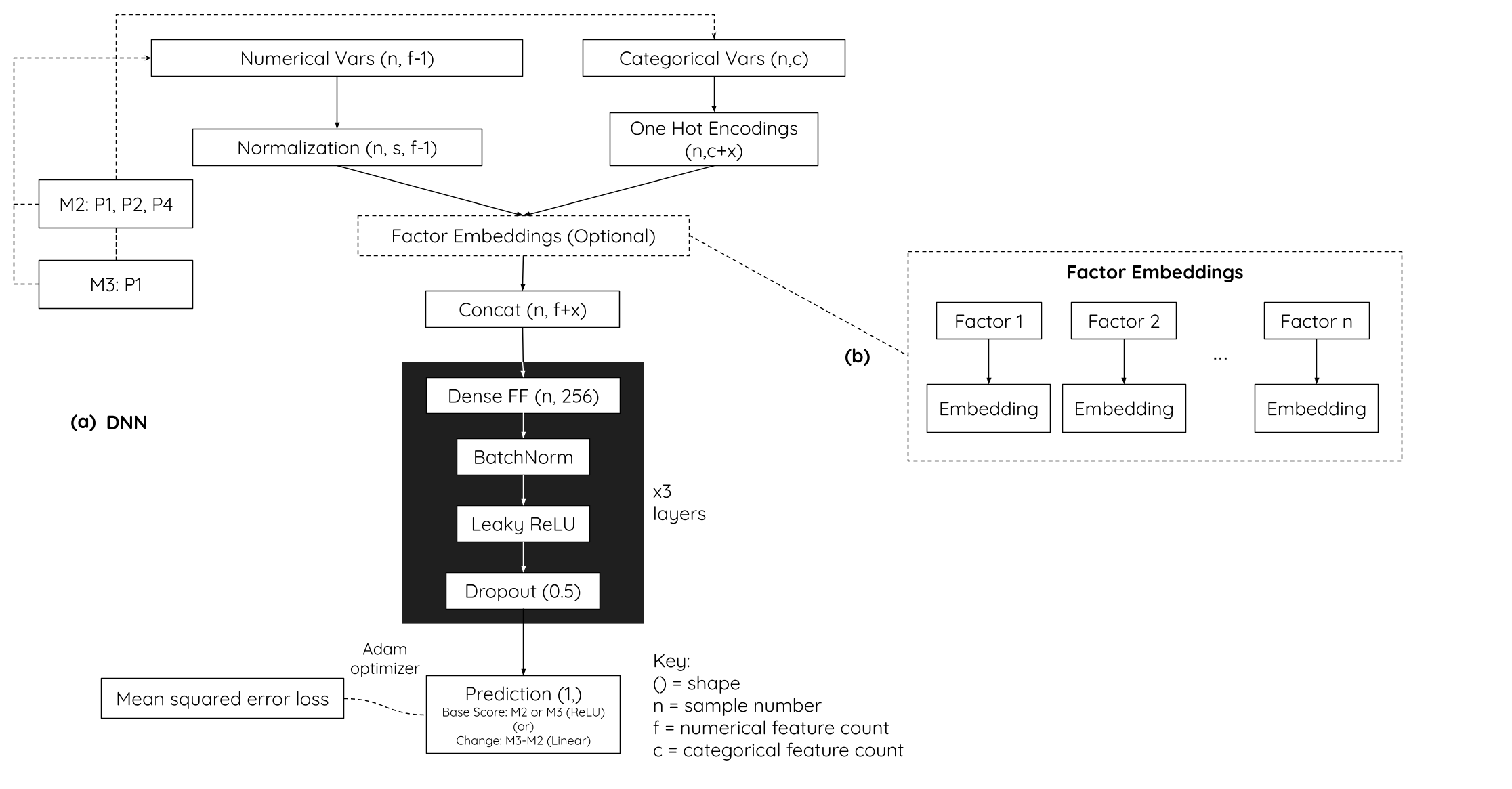}
    \caption{(a) DNN framework. Input data comprises of Survey Measures, Daily Diary and Biomarker data from M2 and Survey Measures data from M3, similar to baseline models. The core architecture comprises of 3 layers of dense networks, with batch normalization, leaky ReLU and a high rate of dropout. The prediction was a ReLU output in cases of positive scores, or linear output when predicting changes, and the network was optimized using a mean squared loss function. (b) In a separate architecture, embeddings were grouped by factors prior to the dense layers and prediction.}
    \label{fig:Deep neural network framework}
\end{figure}

In addition to assessing the performance of off-the-shelf models, we designed a DNN (Figure~\ref{fig:Deep neural network framework}) to provide the expected advantage of non-linear model frameworks, while addressing problems related to overfitting that may be present in a large, sparse dataset like MIDUS-AGG. 

We concatenated the mixed, processed data using hyperparameters optimized through manual search and designed the core architecture that included three layers comprised of: (1) a 256-dimensional dense, feedforward (FF) network, (2) batch normalization, (3) leaky rectified linear unit (Leaky ReLU), and (4) a high dropout rate of 0.5. We utilized a batch size of 32, a ceiling of 10,000 epochs, a learning rate of 0.001 and early stopping loss of 500 epochs. Additionally, we used a random, 15\% of the training data, each epoch, for validation and only saved the DNN weights from the last epoch with an improvement in prediction, through a drop in mean squared error (MSE) loss. We used a cross-validation approach to model fitting as described previously, reporting the average performance of each model.

Base cognitive measures (always $\geq$ 0) were predicted using a ReLU output and change in cognitive scores ($\Delta$) were predicted using a linear output to account for negative values. 

Using an independent split of data, we initially experimented with variants of the model using a standard ReLU, lower (more typical) dropout rates, lower learning rates, and 32, 64, 128 and 512-dimensional FF networks. We experienced common problems faced in machine learning such as vanishing gradients \cite{Hanin2018Which} and overfitting, and determined hyperparameters that contributed to a relative decrease in error rates across cognitive measures.

To assess the independent contribution of each factor, we added an embedding layer \cite{Rong2014Word2vecExplained} to the DNN after data processing (Figure~\ref{fig:Deep neural network framework}b). For example, 274 "Health" factor variables (Supplementary Table~\ref{tab: var-groupings}) were first reduced to a 16-dimensional, learned, vector which was used as input to a DNN. This process is similar to other dimensionality reduction techniques, such as PCA \cite{doi:10.1162/0899766041732396}, but has the added benefit of learning the context of groups of variables during training.

We utilized a total of 55, 16-dimensional embeddings (one for each factor), decreasing the number of inputs to the model to 880. An embedding captured the learned, numerical representation of each group of variables (Table~\ref{tab: var-groupings}). We chose 16 as the number of dimensions since most factors had at least 16 variables, and language models, where embeddings are used extensively, have found success with 16-dimensional vector representations \cite{bredin2017tristounet,barnard2019predicting,vemulapalli2019compact}. Our expectation was that the modularity in designing embeddings will allow a more thorough understanding about each factor's similarity to other factors and to further understand how important factors are to prediction. We used the Kullback-Leibler divergence between the joint probabilities of the average, 16-dimensional embedding and a 2-dimensional coordinate for ease of plotting using t-SNE.

\subsection{Evaluation}

Using the processed, full MIDUS-AGG data and, separately the 16-dimensional data, we fitted the model parameters using MSE as the loss function and report the predicted root mean squared error (RMSE) for each model. Because our goal was prediction, we utilized cross-validation techniques to estimate out-of-sample performance errors. We averaged the error after model fitting and assessment of 10 different splits of the dataset into training (75\%) and test (25\%) data. 

Using the average RMSE for each factor embedding, we ranked predictions from each factor from most important (low RMSE) to least important (high RMSE) and report the average rank of each factor across all cognitive measures. To quantify the similarity between factors, we isolated weights from each factor's embeddings and used t-SNE \cite{vanderMaaten2008VisualizingT-SNE}, a visualization technique used for high-dimensional data. We note, however, there is a well-known trade-off between temporal coherence and projection reliability with visualizing high-dimensional data. 

\subsection{Significance Tests}

Model performance differences, in RMSE, were assessed using paired, pairwise, T-tests with Holm-Bonferroni correction. To test whether sparsity significantly affected model performance, we used an ANOVA to test for interaction effect between sparsity and model type. 

\section{Results}

After data processing, the total number of participants with usable data ranged from 615 for M3 and change measures (M2 mean age=54.67; 54.4\% female) to 881 for M2 measures (M2 mean age=55.74; 56.6\% female). 

\begin{figure}  
    \centering
    \includegraphics[width=\textwidth]{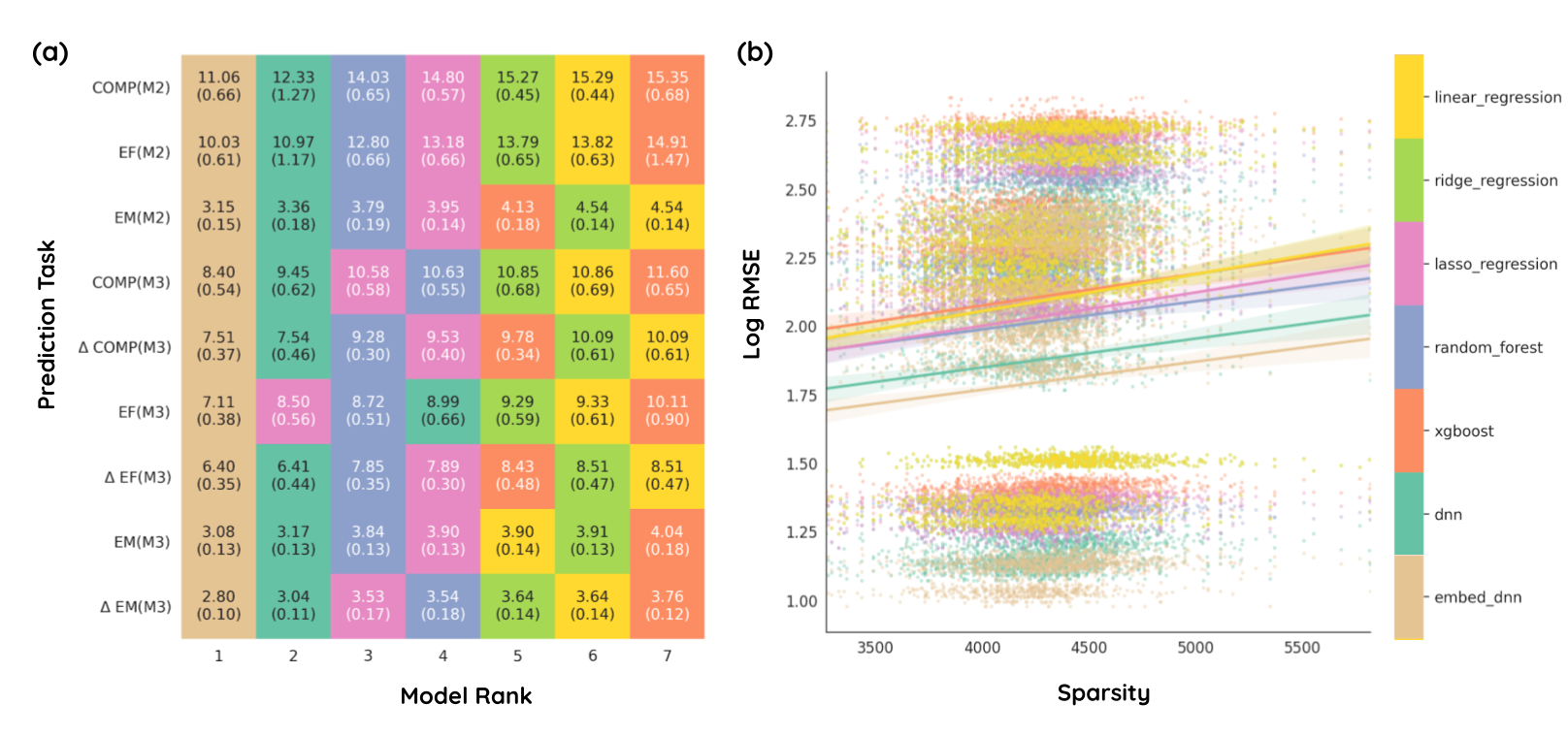}
    \caption{Ranks of cognitive measures. (a) We independently assessed performance from 7 models (colored), as described previously, and ranked the performance using RMSE (lower is better). $embed\_dnn$ refers to the DNN utilizing embeddings from groups of variables, rather than the full dataset as input. SD is displayed in parentheses. (b) Model performance vs. sparsity of input data, averaged across all prediction tasks. Each dot presents a single, test sample, participant, colored by model used in the prediction. A 95\% confidence interval is displayed around each fit.}
    \label{fig:Ranks of cognitive measures}
\end{figure}

\subsection{Model Rankings}
Our primary objective was to assess model performance in predicting executive function and episodic memory, where the predictions were either of raw scores from M2/M3 or change from M2 to M3 ($\Delta$). We ranked performance of models by test data RMSE, averaged over 10, random, splits of the dataset (Figure~\ref{fig:Ranks of cognitive measures}a). This ranking includes the DNN network after reducing the dimensionality of input data through  factor-specific embeddings (i.e., using the $embed\_dnn$ model). 

Across prediction tasks using the full dataset, the DNNs had lower RMSE than other non-linear models (i.e., RF, GB), regularized regression models (i.e., Lasso, Ridge) and unregularized models (i.e., linear regression) (Figure \ref{fig:Ranks of cognitive measures}a) at M2 and M3. There were no significant differences in errors between sex or age groups across all model types ($F_{231,20}=0.80$, p=0.71) (Table ~\ref{rmse descriptive table}). In all 9 tasks, model rankings when predicting M3 measures differed from M2 model rankings. 

We assessed significant differences in errors across model types, through parametric t-test with Holm-Bonferroni correction (Supplementary Table~\ref{tab: Paired ttests}). On average, $embed\_dnn$ significantly differed (T(8)$<$-5.08, p $\leq$ 0.05) in mean RMSE from all other model types, except the DNN. 

\begin{table}
\centering
{
\begin{tabular}{lllrl}
\toprule
Prediction Task & Sex & M2 Age & Sparsity & RMSE (SD)               \\
\midrule
COMP(M2) & Female & $<$55 &   4327.78 &  14.04 (0.41) \\
       &      & $\ge$55 &   4333.53 &  13.99 (0.42) \\
       & Male & $<$55 &   4457.71 &  14.02 (0.44) \\
       &      & $\ge$55 &   4489.13 &  14.04 (0.41) \\\hdashline
EF(M2) & Female & $<$55 &   4320.30 &   12.8 (0.52) \\
       &      & $\ge$55 &   4332.89 &  12.78 (0.53) \\
       & Male & $<$55 &   4458.73 &   12.8 (0.52) \\
       &      & $\ge$55 &   4482.27 &  12.74 (0.51) \\\hdashline
EM(M2) & Female & $<$55 &   4331.18 &    3.92 (0.1) \\
       &      & $\ge$55 &   4338.65 &    3.92 (0.1) \\
       & Male & $<$55 &   4468.83 &    3.92 (0.1) \\
       &      & $\ge$55 &   4492.75 &    3.93 (0.1) \\\hdashline
COMP(M3) & Female & $<$55 &   4104.33 &  10.34 (0.39) \\
       &      & $\ge$55 &   4137.47 &  10.33 (0.39) \\
       & Male & $<$55 &   4304.29 &  10.33 (0.39) \\
       &      & $\ge$55 &   4274.46 &   10.34 (0.4) \\\hdashline
$\Delta$ COMP(M3) & Female & $<$55 &   4110.12 &   9.12 (0.26) \\
       &      & $\ge$55 &   4137.43 &    9.1 (0.26) \\
       & Male & $<$55 &   4287.56 &   9.13 (0.28) \\
       &      & $\ge$55 &   4263.45 &    9.1 (0.27) \\\hdashline
EF(M3) & Female & $<$55 &   4101.07 &   8.87 (0.37) \\
       &      & $\ge$55 &   4122.14 &    8.86 (0.4) \\
       & Male & $<$55 &   4304.95 &   8.86 (0.38) \\
       &      & $\ge$55 &   4274.40 &   8.89 (0.39) \\\hdashline
$\Delta$ EF(M3) & Female & $<$55 &   4105.48 &   7.72 (0.28) \\
       &      & $\ge$55 &   4125.95 &   7.72 (0.25) \\
       & Male & $<$55 &   4298.70 &   7.72 (0.25) \\
       &      & $\ge$55 &   4270.60 &    7.7 (0.24) \\\hdashline
EM(M3) & Female & $<$55 &   4104.97 &   3.69 (0.08) \\
       &      & $\ge$55 &   4137.07 &   3.69 (0.08) \\
       & Male & $<$55 &   4297.32 &   3.69 (0.09) \\
       &      & $\ge$55 &   4290.35 &    3.69 (0.1) \\\hdashline
$\Delta$ EM(M3) & Female & $<$55 &   4119.27 &   3.42 (0.09) \\
       &      & $\ge$55 &   4118.28 &   3.42 (0.09) \\
       & Male & $<$55 &   4286.56 &   3.42 (0.08) \\
       &      & $\ge$55 &   4273.48 &   3.43 (0.09) \\
\bottomrule
\end{tabular}}
\caption{Breakdown of RMSE by sex and average M2 age. Sparsity is averaged by subject, and represents the amount of missing predictor variables. RMSE is averaged across all models. There was no significant difference in RMSE between sex and ages.}
\label{rmse descriptive table}
\end{table}

\subsection{Sparsity Effect}

Figure \ref{fig:Ranks of cognitive measures}b shows the effect of sparsity on predicting cognitive performance across all tasks. The slopes of the linear trends appear different for DNN models compared to the other model types, suggesting a possible interaction effect of sparsity on model performance. We tested the interaction effect of sparsity and model type on RMSE with a ANOVA for two linear regression models: without a sparsity interaction term ($F_{36239,6} = 327.10$,p $<$ 0.01) and with a sparsity interaction term ($F_{36232,13} = 184.50$,p $<$ 0.01) on model type. We found a significant interaction effect ($F_{9} = 59.20$,p $<$ 0.01), indicating that DNN models are affected less by sparsity than the other models we assessed.

\begin{figure}
    \centering
    \includegraphics[width=\textwidth]{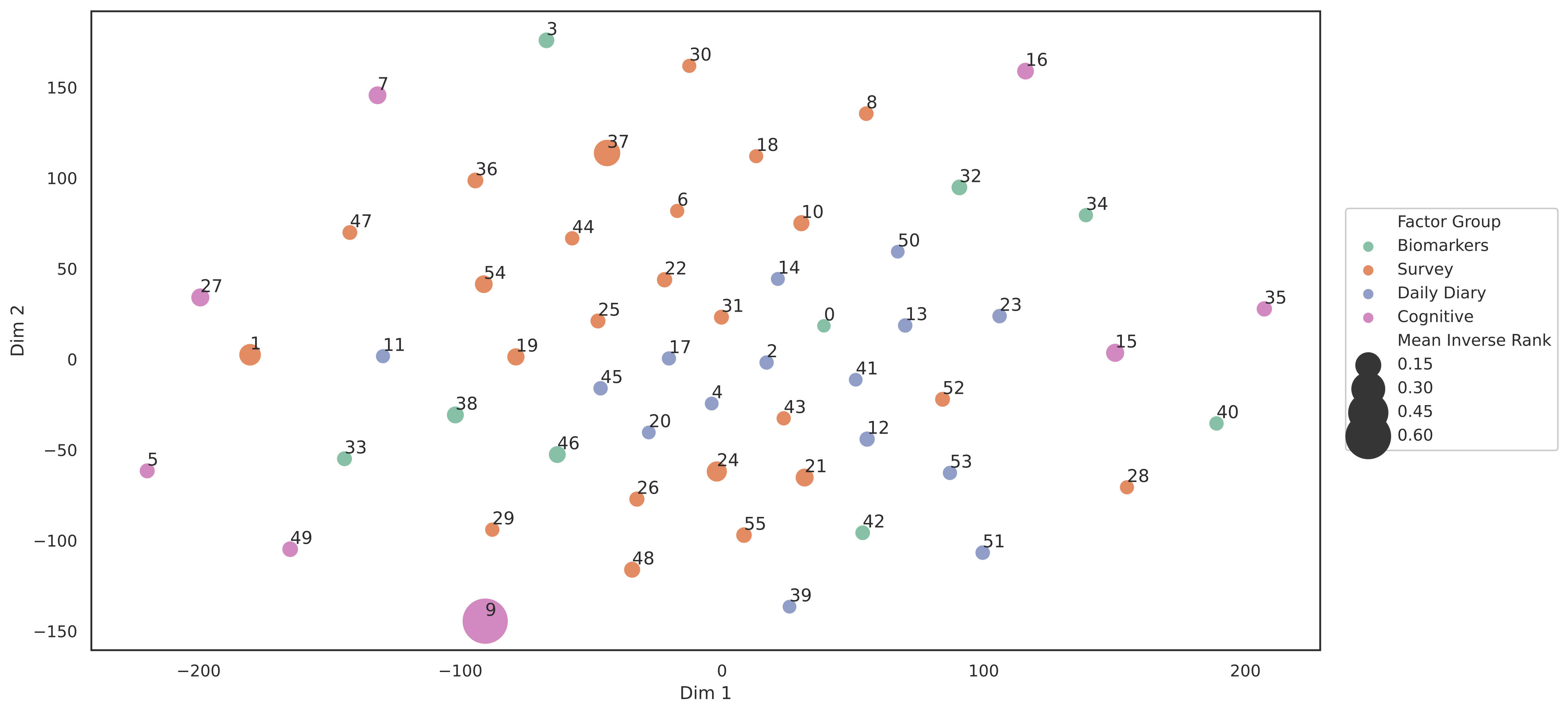}
    \caption{2-dimensional, t-SNE, visualization representing learned embeddings of factor similarity. Each dot represents  a 16-dimensional, factor embedding learned during training and used as DNN input for prediction of test data. We display the index of factors, and detail the corresponding factors and measures in Table~\ref{tab: embedding rankings}. Each factor was independently used for prediction, and we averaged across prediction tasks. The inverse rank of each factor (i.e., its importance) is represented by the diameter of the circle. Each color represents the hierarchical category, and distance represents the relative relatedness of each factor to others. Note that the cognitive factors (pink) were only used when predicting (M3) and $\Delta$ cognitive measures.}
    \label{fig:embedding-vis}
\end{figure}

\begin{table}
\centering
\resizebox{.9\textwidth}{!}{
\begin{tabular}{lllll}
\toprule
Index & Factor & Project & Average Rank & Mean Inverse Rank\\
\midrule
     0 &                                  Actigraphy &   Biomarkers &         45.56 &               0.02 \\
     1 &                              Administration &       Survey &          8.11 &               0.12 \\
     2 &                                      Affect &  Daily Diary &         32.67 &               0.03 \\
     3 &                                  Assay Data &   Biomarkers &         22.11 &               0.05 \\
     4 &                                  Assistance &  Daily Diary &         39.44 &               0.03 \\
     5 &                           Backward Counting &    Cognitive &         26.67 &               0.04 \\
     6 &                                  Caregiving &       Survey &         34.56 &               0.03 \\
     7 &                            Category Fluency &    Cognitive &         14.17 &               0.07 \\
     8 &                                    Children &       Survey &         29.22 &               0.03 \\
     9 &             Cognitive Battery Factor Scores &    Cognitive &          1.50 &               0.67 \\\hdashline
    10 &                       Community Involvement &       Survey &         20.00 &               0.05 \\
    11 &                                    Cortisol &  Daily Diary &         35.44 &               0.03 \\
    12 &                        Daily Discrimination &  Daily Diary &         24.67 &               0.04 \\
    13 &                           Daily Medications &  Daily Diary &         31.44 &               0.03 \\
    14 &                             Daily Stressors &  Daily Diary &         37.33 &               0.03 \\
    15 &                    Delayed Word List Recall &    Cognitive &         13.50 &               0.07 \\
    16 &                             Digits Backward &    Cognitive &         17.33 &               0.06 \\
    17 &                       Disability Assistance &  Daily Diary &         33.44 &               0.03 \\
    18 &                              Discrimination &       Survey &         35.22 &               0.03 \\
    19 &   Education, Occupation, and Marital Status &       Survey &         15.78 &               0.06 \\\hdashline
    20 &                           Emotional Support &  Daily Diary &         39.89 &               0.03 \\
    21 &                                    Finances &       Survey &         13.78 &               0.07 \\
    22 &                                      Health &       Survey &         24.11 &               0.04 \\
    23 &                            Health Behaviors &  Daily Diary &         31.22 &               0.03 \\
    24 &                            Health Insurance &       Survey &          9.78 &               0.10 \\
    25 &                  Health Questions for Women &       Survey &         26.89 &               0.04 \\
    26 &               Household Roster and Children &       Survey &         25.22 &               0.04 \\
    27 &                  Immediate Word List Recall &    Cognitive &         13.67 &               0.07 \\
    28 &                                Life Overall &       Survey &         35.00 &               0.03 \\
    29 &                           Life Satisfaction &       Survey &         33.00 &               0.03 \\\hdashline
    30 &                         Living Arrangements &       Survey &         35.00 &               0.03 \\
    31 &              Marriage or Close Relationship &       Survey &         26.67 &               0.04 \\
    32 &                             Medical History &   Biomarkers &         21.89 &               0.05 \\
    33 &                            Medication Chart &   Biomarkers &         27.56 &               0.04 \\
    34 &                             Musculoskeletal &   Biomarkers &         33.33 &               0.03 \\
    35 &                               Number Series &    Cognitive &         24.17 &               0.04 \\
    36 &                             Parent's Health &       Survey &         21.44 &               0.05 \\
    37 &                            Personal Beliefs &       Survey &          4.89 &               0.20 \\
    38 &                               Physical Exam &   Biomarkers &         16.78 &               0.06 \\
    39 &                           Physical Symptoms &  Daily Diary &         40.22 &               0.02 \\\hdashline
    40 &        Pittsburgh Sleep Questionnaire (PSQ) &   Biomarkers &         32.11 &               0.03 \\
    41 &                             Positive Events &  Daily Diary &         40.78 &               0.02 \\
    42 &                   Psychophysiology Protocol &   Biomarkers &         29.22 &               0.03 \\
    43 &                          Race and Ethnicity &       Survey &         33.22 &               0.03 \\
    44 &                   Religion and Spirituality &       Survey &         32.22 &               0.03 \\
    45 &                             Scale Variables &  Daily Diary &         31.78 &               0.03 \\
    46 &      Self-Administrated Questionnaire (SAQ) &   Biomarkers &         17.00 &               0.06 \\
    47 &                                   Sexuality &       Survey &         28.00 &               0.04 \\
    48 &                             Social Networks &       Survey &         20.33 &               0.05 \\
    49 &  Stop and Go Switch Task - Composite Scores &    Cognitive &         22.33 &               0.04 \\\hdashline
    50 &                                    Time use &  Daily Diary &         41.11 &               0.02 \\
    51 &                                Week Summary &  Daily Diary &         30.67 &               0.03 \\
    52 &                                        Work &       Survey &         27.33 &               0.04 \\
    53 &                              Work Behaviors &  Daily Diary &         33.56 &               0.03 \\
    54 &                                 Your Health &       Survey &         14.44 &               0.07 \\
    55 &                           Your Neighborhood &       Survey &         22.78 &               0.04 \\
\bottomrule
\end{tabular}}
\caption{Embedding similarity and ranking. The diameter of the points in Figure \ref{fig:embedding-vis}, i.e., importance towards prediction, was calculated as the inverse of the average rank across variables.} 
\label{tab: embedding rankings}
\end{table}

\subsection{Embedding Similarity}

During training for each prediction task, the $embed\_dnn$ model learned the 16-dimensional embedding of related variables, i.e., factors. Using t-SNE to visualize the factor embeddings, we show each factor on a reduced, 2-dimensional plot, colored by its hierarchical group, using labels specified by the MIDUS project (Figure ~\ref{fig:embedding-vis}). We sized each factor in the plot by its independent influence on decreasing RMSE. Cognitive measures were only used in the prediction of raw and $\Delta$ M3 measures. In other words, we used M2 cognitive measures as variables when predicting M3 cognitive performance. No cognitive variables were used in cross-sectional prediction of M2 cognitive measures. Corresponding labels of indices and average rank measures are detailed in Table~\ref{tab: embedding rankings}.

We noted some clustering of cognitive measures in their embedding representation, and their relatively large influence when used to predict future cognitive performance. On average, the most important factors were "Administrative" or demographic in nature (i.e., age and gender), related to education, occupation, marital status, or self-described psychological well-being and subjective age ("Personal Beliefs"). Questions related to menopause, and hormonal drugs were also important in predicting cognitive performance for women. Additionally, questions related to participants' health insurance status (e.g., whether they were covered by private health insurance or Medicare) were important in prediction. While questions related to medical history and biomarker data (e.g., "Cortisol") possibly improved prediction differently than other factors containing variables describing medications, medical history, psychophysiology, biomarker sleep quality and physical exams generally ranked lower. Specifically, the top five factors that most improved prediction (average rank $\leq$ 13.5), in order, were (1) Cognitive Battery Factor Scores, (2) Personal Beliefs, (3) Administration, (4) Health Insurance, (5) Delayed Word List Recall. The bottom five factors that least improved prediction (average rank $\geq$ 39.5), in order, were (1) Actigraphy, (2) Time use, (3) Positive Events, (4) Physical Symptoms, (5) Emotional Support.  

Interestingly, cognitive factors such as the "Immediate Word List Recall" and "Stop and Go Switch Task" measures did not clearly cluster together. This may indicate that the DNN models interpreted these factors as independently contributing to better prediction of cognitive performance relative to other cognitive measures.

\section{Discussion}

In this study, we developed a modular approach to improving prediction of cognitive performance in a sparse, heterogeneous dataset, which may be used to diagnose conditions that rely on accurate executive function and episodic memory measurements, such as ADHD, MCI and AD. By using a large dataset containing behavioral, biological and psychosocial data, we reported the relative advantage of deep neural networks, compared to linear and other non-linear models, in consistently improving prediction of cognitive scores, under varying levels of data sparsity. Through evaluating cognitive performance prediction using over 5000 predictor variables, across 9 prediction tasks, we found the DNN framework produced the most accurate and consistent results. Our cross-validation approach to measure mean performance and variance of errors across frameworks did not produce significant differences between sex and age groups.

The DNN performance yielded lower RMSEs in all prediction tasks compared to other models (Figure~\ref{fig:Ranks of cognitive measures}a). Because prediction of change in cognitive performance is often a more challenging problem, deep architectures may be preferable to other frameworks, as they are more likely to model the complicated, temporal relationships between behavioral, physiological and other factors that affect cognitive decline. 

The consistency in the rankings of DNNs over other model types we tested may be due to their relative advantage under data sparsity. By showing a significant interaction between model type and sparsity in prediction cognitive performance, we show that deep networks may be less affected by varying levels of sparsity than linear or other non-linear models. This is visualized by the relatively flatter slopes of DNN and $embed\_dnn$ in Figure ~\ref{fig:Ranks of cognitive measures}b. The methods we employed to control for overfitting may have allowed better generalization and robustness of DNNs, relative to other model types we assessed.

While we expected factor embeddings to outperform linear methods, we were surprised to find that, for all of the assessed prediction tasks, using embeddings yielded more accurate predictions than the DNN with access to the full dataset on average, although this finding not statistically significant. However, we achieved similar accuracy despite a tenfold reduction to the number of input variables. 

By providing the DNN with relevant information in how similar variables are to each other, we allowed meaningful context when predicting cognitive performance, and achieved better prediction accuracy with fewer parameters and in fewer iterations. This finding is similar to other observations for DNNs, where hyperparameter design choice with the dataset in mind both improves accuracy and training time \cite{smith2017cyclical}. Expert clinical guidance on how variables are grouped prior to prediction in a deep network may thus improve performance in and interpretability of similar datasets. 

We believe our pattern of results is related to the sparseness and variability present in the full MIDUS-AGG dataset, which is less of an issue when first grouping variables before training. In fact, architectures designed with characteristics of sparse data in mind during design tend to optimize faster and avoid local minima \cite{Duchi2013EstimationSparse}. We hypothesize that, given a low enough learning rate and long enough training time, our DNN using the full dataset would eventually outperform the DNN using embeddings as input. Our DNN performance may also be improved by inclusion of residual connections that may yield loss functions that are easier to minimize \cite{Li2018VisualizingNets}. 

We recommend future study to understand how the size and sparseness of clinical data may influence the rate of learning, which may have resulted in this finding. Prediction using meaningful groupings of variables is of interest to clinicians and researchers who work with large, sparse datasets like MIDUS-AGG. Working with embeddings allows prediction models to scale efficiently as dataset sizes grow in volume. Through improving prediction of cognitive performance, clinicians and researchers, often constrained to working with sparse data, may be better able to diagnose a host of clinical conditions that rely on accurate measurement of cognitive performance.

\section{Conclusion}

In this evaluation of deep neural architectures to improve prediction of cognitive performance, we found robust performance of DNNs under multiple, cross-sectional and longitudinal, measures. We show that the robustness can partly be attributed to how DNNs handle data sparsity compared with linear and other non-linear models. Using factor-level, low-dimensional embeddings, we find no drop in model accuracy while allowing clinical interpretability of variables impacting executive and episodic function. Our results highlight the potential of DNNs to improve clinical understanding of age-related cognitive decline and in the analysis of large, mixed datasets.

\section*{Acknowledgement}
This work was supported by grants from the National Science Foundation (IIS-1513853, IIS-1816363), the Army Research Laboratory Cooperative Agreement W911NF-10-2-0022, a Vannevar Bush Faculty Fellowship from the US Department of Defense (N00014-20-1-2027), and National Institute of Aging (P01-AG020166 and U19-AG051426). 

\printbibliography

\newpage
\section*{Supplementary material}

\begin{table}[ht]
\centering
\resizebox{!}{.6\height}{
\begin{tabular}{clrrr} 
\toprule
 Project & Factors & Categorical Variables & Numerical Variables & Total Variables \\ 
\midrule

 & Administration & 3 & 3 & 6 \\
 & Caregiving & 20 & 5 & 25 \\
 & Children & 16 & 8 & 24 \\
 & Community Involvement & 30 & 59 & 89 \\
 & Discrimination & 21 & 13 & 34 \\
 & Education, Occupation, and Marital Status & 71 & 29 & 100 \\
 & Finances & 34 & 35 & 69 \\
 & Health & 231 & 43 & 274 \\
 & Health Insurance & 42 & \multicolumn{1}{l}{} & 42 \\
 & Health Questions for Women & 53 & 4 & 57 \\
 \multirow{0}{*}{Survey} & Household Roster and Children & 256 & 68 & 324 \\
 & Life Overall & \multicolumn{1}{l}{} & 6 & 6 \\
 & Life Satisfaction & 10 & \multicolumn{1}{l}{} & 10 \\
 & Living Arrangements & 11 & 2 & 13 \\
 & Marriage or Close Relationship & 36 & 21 & 57 \\
 & Parent's Health & 6 & 4 & 10 \\
 & Personal Beliefs & 310 & 95 & 405 \\
 & Race and Ethnicity & 71 & \multicolumn{1}{l}{} & 71 \\
 & Religion and Spirituality & 48 & 8 & 56 \\
 & Sexuality & 6 & 6 & 12 \\
 & Social Networks & 52 & 7 & 59 \\
 & Work & 108 & 31 & 139 \\
 & Your Health (Self Evaluation) & 294 & 55 & 349 \\
 & Your Neighborhood (Self Evaluation) & 16 & 4 & 20 \\  

  & \multicolumn{1}{c}{Group Total} & 1745 & 506 & 2251 \\ 
\cline{2-5} 
\multirow{3}{*}{Cognitive} \\
  & BTACT & \multicolumn{1}{l}{} & 10 & 10 \\
  & \multicolumn{1}{c}{Group Total} & \multicolumn{1}{c}{~} & 10 & 10 \\
\cline{2-5} 
 \multirow{16}{*}{Daily Diary} \\
 & Affect & 27 & \multicolumn{1}{l}{} & 27 \\
 & Assistance & 52 & 4 & 56 \\
 & Cortisol & 4 & 4 & 8 \\
 & Daily Discrimination & 21 & \multicolumn{1}{l}{} & 21 \\
 & Daily Medications & 10 & \multicolumn{1}{l}{} & 10 \\
 & Daily Stressors & 129 & 14 & 143 \\
 & Disability Assistance & 26 & 2 & 28 \\
 & Emotional Support & 51 & 4 & 55 \\
 & Health Behaviors & 2 & \multicolumn{1}{l}{} & 2 \\
 & Physical Symptoms & 56 & \multicolumn{1}{l}{} & 56 \\
 & Positive Events & 20 & 10 & 30 \\
 & Scale Variables & 13 & 9 & 22 \\
 & Time use & 13 & 16 & 29 \\
 & Week Summary & 35 & \multicolumn{1}{l}{} & 35 \\
 & Work Behaviors & 8 & \multicolumn{1}{l}{} & 8 \\ 
 & \multicolumn{1}{c}{Group Total} & 467 & 63 & 530 \\
\cline{2-5}
\multirow{10}{*}{Biomarkers} \\
 & Actigraphy & 128 & 217 & 345 \\
 & Assay Data & 1 & 62 & 63 \\
 & Medical History & 493 & 94 & 587 \\
 & Medication Chart & 204 & 81 & 285 \\
 & Musculoskeletal & 6 & 17 & 23 \\
 & Physical Exam & 263 & 42 & 305 \\
 & Pittsburgh Sleep Questionnaire (PSQ) & 22 & 10 & 32 \\
 & Psychophysiology Protocol & 32 & 153 & 185 \\
 & Self-Administrated Questionnaire (SAQ) & 405 & 43 & 448 \\ 
  & \multicolumn{1}{c}{Group Total} & 1554 & 719 & 2273 \\ 
\cline{2-5}
\\
\multicolumn{2}{c}{Total Counts} & 3766 & 1298 & 5064 \\
\bottomrule
\end{tabular}
}
\caption{Variables used in the present study, by groups. Groups were utilized for factor analysis and embeddings in the DNN.}
\label{tab: var-groupings}
% COMMENT: what are the sample sizes for each of the categories of variables
\end{table}

\begin{table*}
\centering
{\begin{tabular}{lllll@{}}
\toprule
 Model A & \multicolumn{1}{c}{Model B} & \multicolumn{1}{c}{T} & \multicolumn{1}{c}{dof} & \multicolumn{1}{c}{p-corrected}\\ \midrule
embed\_{dnn}         & dnn                & 2.83  & 8   & 0.11        \\
embed\_{dnn}         & lasso\_regression  & -5.08 & 8   & 0.02*        \\
embed\_{dnn}         & xgboost            & -5.17 & 8   & 0.02*        \\
embed\_{dnn}         & linear\_regression & -5.75 & 8   & 0.01**        \\
embed\_{dnn}         & random\_forest     & -5.74 & 8   & 0.01**        \\
embed\_{dnn}         & ridge\_regression  & -5.77 & 8   & 0.01**        \\ \hdashline
dnn                & lasso\_regression  & -3.70 & 8   & 0.04*        \\
dnn                & random\_forest     & -4.23 & 8   & 0.03*        \\
dnn                & ridge\_regression  & -4.86 & 8   & 0.02*        \\
dnn                & linear\_regression & -4.88 & 8   & 0.02*        \\
dnn                & xgboost            & -5.04 & 8   & 0.02*        \\ \hdashline
ridge\_regression  & lasso\_regression  & 5.14  & 8   & 0.02*        \\
ridge\_regression  & random\_forest     & 4.45  & 8   & 0.02*        \\
ridge\_regression  & xgboost            & -1.40 & 8   & 0.45        \\
ridge\_regression  & linear\_regression & -2.43 & 8   & 0.17        \\ \hdashline
linear\_regression & lasso\_regression  & 5.12  & 8   & 0.02*        \\
linear\_regression & random\_forest     & 4.45  & 8   & 0.02*        \\ 
linear\_regression & xgboost            & -1.36 & 8   & 0.45        \\ \hdashline
lasso\_regression  & random\_forest     & 1.59  & 8   & 0.45        \\
lasso\_regression  & xgboost            & -3.37 & 8   & 0.06        \\ \hdashline
random\_forest     & xgboost            & -3.90 & 8   & 0.04* \\
\bottomrule
\end{tabular}}
\caption{Paired T-Tests: Pairwise, parametric, T-tests between model types. The Holm-Bonferroni correction was used to judge significance. \\
\ * $\leq$ 0.05, ** $\leq$ 0.01} 
\label{tab: Paired ttests}
\end{table*}

\end{document}